\pdfoutput=1

\documentclass[11pt]{article}

\usepackage[preprint]{acl}

\usepackage{times}
\usepackage{latexsym}

\usepackage[T1]{fontenc}

\usepackage[utf8]{inputenc}

\usepackage{microtype}

\usepackage{inconsolata}

\usepackage{graphicx}

\usepackage{color-edits}
\addauthor[Alexis]{ap}{cyan}
\addauthor[Michael]{mg}{red}
\usepackage{gb4e}
\noautomath

\usepackage{amsmath}

\usepackage{subcaption}
\usepackage{listings}
\usepackage{xcolor}
\usepackage{booktabs}
\usepackage{multicol}
\usepackage{multirow}

\title{Can we teach language models to gloss endangered languages?}

\author{Michael Ginn${}^{1}$ \and Mans Hulden${}^{2}$ \and Alexis Palmer${}^{1}$ \\
    ${}^{1}$University of Colorado \quad ${}^{2}$New College of Florida \\
  \texttt{michael.ginn@colorado.edu} \\}

\begin{document}
\maketitle
\begin{abstract}
Interlinear glossed text (IGT) is a popular format in language documentation projects, where each morpheme is labeled with a descriptive annotation. Automating the creation of interlinear glossed text would be desirable to reduce annotator effort and maintain consistency across annotated corpora. Prior research \citep{ginn-etal-2023-findings, zhao-etal-2020-automatic, moeller-hulden-2018-automatic} has explored a number of statistical and neural methods for automatically producing IGT. 

As large language models (LLMs) have showed promising results across multilingual tasks, even for rare, endangered languages \citep{zhang2024hire}, it is natural to wonder whether they can be utilized for the task of generating IGT.  We explore whether LLMs can be effective at the task of interlinear glossing with in-context learning, without any traditional training. We propose new approaches for selecting examples to provide in-context, observing that targeted selection can significantly improve performance. We find that LLM-based methods beat standard transformer baselines, despite requiring no training at all.
These approaches still underperform state-of-the-art supervised systems for the task, but are highly practical for researchers outside of the NLP community, requiring minimal effort to use.

\end{abstract}

\section{Introduction}
With thousands of endangered languages at risk of extinction, language documentation has become a major area of linguistic research \citep{himmelmann2006language, woodbury1999language}, aiming to produce permanent artifacts such as annotated corpora, reference grammars,
and dictionaries. Furthermore, research has explored the potential for computational methods to aid in language documentation and revitalization \citep{palmer-etal-2009-evaluating, moeller-hulden-2018-automatic, wiemerslage-etal-2022-morphological, kann-etal-2022-machine, gessler-2022-closing, zariquiey-etal-2022-cld2,zhang-etal-2022-nlp, flavelle-lachler-2023-strengthening}. 


\begin{figure}
    \centering
    \includegraphics[width=\linewidth]{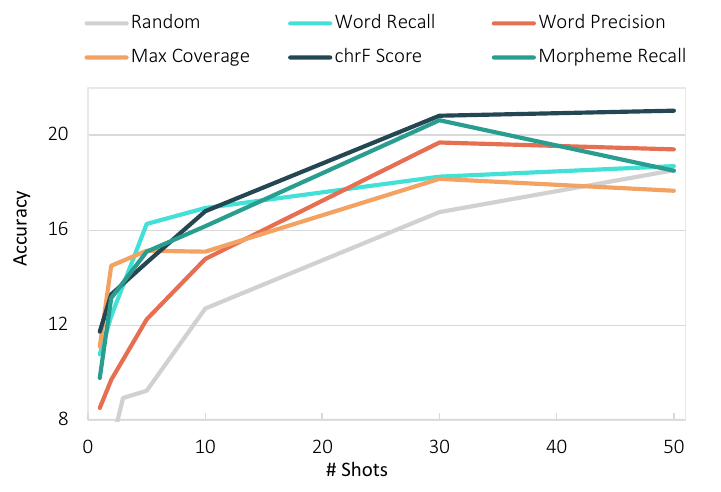}
    \caption{Accuracy of an LLM-based glossing method on Gitksan data, varying the number of provided examples and the strategy for selecting examples.}
    \label{fig:gitx-shots}
\end{figure}

In particular, we study the task of generating interlinear glossed text (IGT), a line-by-line format for annotated text corpora that is commonly used in documentation projects. IGT generation has been studied using statistical \citep{palmer-etal-2009-evaluating, samardzic-etal-2015-automatic, mcmillan-major-2020-automating} and neural \citep{moeller-hulden-2018-automatic, zhao-etal-2020-automatic, barriga-martinez-etal-2021-automatic} methods.

A key challenge when working with endangered languages is that, in nearly all cases,\footnote{As \citet{liu-etal-2022-always} notes, not all endangered languages are low-resource (and vice versa), and such languages bear different concerns when developing language technology.} there is very little labeled or unlabeled data available. This is particularly challenging for large neural models which depend on large, representative training data sets. Research has explored methods to overcome this challenge for IGT generation systems, such as crosslingual transfer \citep{he-etal-2023-sigmorefun, okabe-yvon-2023-towards, ginn2024glosslm} and architectural modifications \citep{girrbach-2023-tu}, but these approaches struggle in very low-resource scenarios.
In addition, previous approaches generally require expertise in model training, implementation, and deployment, as well as  the computational resources needed to serve large neural models.

As large language models (LLMs) have demonstrated impressive performance on various natural language tasks, the question arises whether they can benefit language documentation. We seek to evaluate the ability of current LLMs to generate interlinear glossed text, compared with earlier state-of-the-art methods. This research can also shed light on the language-agnostic capabilities of LLMs, requiring the model to learn patterns in very rare languages which 
are unlikely to have significant presence in their training data. 

We study strategies for selecting in-context examples, finding significant impacts to performance. Our best-performing systems outperform transformer model baselines, despite involving no training whatsoever. They still underperform SOTA systems that induce morphological segmentation, but at the same time hold promise for offering a new approach to interlinear glossing for language documentation practitioners. Our code is available on Github.\footnote{\url{https://github.com/michaelpginn/igt-icl}}


\section{Background}
\subsection{Interlinear Glosed Text}
A typical example of IGT is shown in \autoref{ex:arapaho-gloss}.

\begin{small}
\begin{exe}
  \ex
  \gll nuhu' tih-'eeneti-3i' heneenei3oobei-3i' \\
       this when.PAST-speak-3PL IC.tell.the.truth-3PL \\
  \glt ``When they speak, they tell the truth.'' \citep{cowell2020}
  \label{ex:arapaho-gloss}
\end{exe}
\end{small}

The first line (transcription line) contains the text in the language being documented, and may be segmented into morphemes (as here). The second line (gloss line) provides a \textit{gloss} for each morpheme in the transcription. Glosses may indicate grammatical function or a translation of the morpheme (for stems). The third line  contains a translation into a high-resource language such as English. Producing each of these lines requires knowledge of the language and/or skilled linguistic analysis.

Generally, automated IGT systems are trained to predict the gloss line given the transcription line (and sometimes the translation as in \citealp{zhao-etal-2020-automatic, rice2024tams}). The primary aim of such systems is to assist a human annotator, providing suggestions for common morphemes that are often glossed with the same label. These systems are not intended to replace human annotators, who are vital to the documentation process, annotating novel morphemes and interesting linguistic phenomena, as well as verifying automatically-produced labels.

\subsection{LLMs for Rare Languages}
Though LLMs generally have limited understanding of rare and low-resource languages \citep{ebrahimi-etal-2022-americasnli}, they can often achieve significantly better performance through \textbf{crosslingual in-context learning} (X-ICL), where a number of examples in the target language are provided directly in the prompt to a multilingual model \citep{winata-etal-2021-language, lin-etal-2022-shot, cahyawijaya2024llms}.

We study X-ICL methods for using LLMs for the task of IGT generation, including complete IGT examples in the prompt. We hypothesize that this approach will leverage both the set of labeled training examples and the robust multilingual knowledge of the language model. In particular, we explore the effects of including an increasing number of examples in context (\autoref{sec:num-shots}) and using different strategies to select relevant examples (\autoref{sec:retrieval}). 


\subsection{Related Work}
A number of approaches have been used for IGT generation. \citet{palmer-etal-2009-evaluating} uses a maximum entropy classifier and represents the earliest work describing benefits of using automated glossing systems. A number of papers \citep{samardzic-etal-2015-automatic, moeller-hulden-2018-automatic, mcmillan-major-2020-automating} use statistical classifiers such as conditional random fields. Recent research explores neural models such as recurrent neural networks and transformers \citep{moeller-hulden-2018-automatic, zhao-etal-2020-automatic, barriga-martinez-etal-2021-automatic}. Other approaches improve glossing performance using crosslingual transfer \citep{he-etal-2023-sigmorefun, okabe-yvon-2023-towards, ginn2024glosslm}, hard attention \citep{girrbach-2023-tu}, and pseudolabeling \citep{ginn-palmer-2023-robust}.

IGT data is not only useful for preservation and revitalization projects, but also for downstream tasks such as machine translation \citep{zhou2019using}, developing linguistic resources like dictionaries \citep{beermann-etal-2020-developing} and UMR (Uniform Meaning Representation) graphs \citep{buchholz-etal-2024-bootstrapping-umr}, studying syntax and morphology \citep{bender-etal-2013-towards, zamaraeva-2016-inferring, moeller-etal-2020-igt2p}, and dependency parsing \citep{georgi-etal-2012-improving}.

Given the cost and difficulty of obtaining IGT data, research has explored methods to scrape it from \LaTeX~ documents \citep{schenner-nordhoff-2016-extracting, nordhoff-kramer-2022-imtvault} and even images \citep{round-etal-2020-automated}. Finally, another line of work has attempted to standardize IGT conventions and formats, balancing consistency and expressiveness across languages \citep{lehmann1982, hughes-etal-2003-encoding, nordhoff-2020-modelling, mortensen-etal-2023-generalized}.
 
\section{Methodology}
We study the IGT generation task described in \citet{ginn-etal-2023-findings}. Given a transcription line and translation line, systems must predict the gloss line. We focus on the \textit{closed track} setting, where the input words are not segmented into morphemes. This task is strictly more difficult than the setting where words are already segmented, as models must jointly learn segmentation and gloss prediction. As reported in \citet{ginn-etal-2023-findings}, the SOTA on this task remains far weaker than the setting with segmented inputs, with up to a 40 point discrepency in SOTA performance. 

\subsection{Data}
We use the IGT corpora and splits from the 2023 SIGMORPHON Shared Task \citep{ginn-etal-2023-findings}, allowing us to directly compare several other systems. We use the languages described in \autoref{tab:data}.

\begin{table}[htb]
    \centering
    \begin{tabular}{l|c c c}
        \hline
         & \multicolumn{3}{c}{\# IGT Examples} \\
        Language & Train & Dev & Test \\
        \hline
        Gitskan [git] & 74 & 42 & 31 \\
        Lezgi [lez] & 705 & 88 & 87 \\
        Natugu [ntu] & 791 & 99 & 99 \\
        Uspanteko [usp] & 9774 & 232 & 633 \\
        \hline
    \end{tabular}
    \caption{Languages and data splits, originally from \citet{ginn-etal-2023-findings}}
    \label{tab:data}
\end{table}

We primarily focus on the lower-resource languages from the shared task, where neural methods tended to struggle due to limited training data. We use the data as formatted by \citet{ginn2024glosslm}.

\subsection{Evaluation}
We evaluate using the same metrics as the shared task. We primarily report \textit{morpheme accuracy}, which measures how many morpheme glosses match between the predicted and true glosses. Any predicted glosses beyond the length of the true gloss string are ignored.

\subsection{Models}
We run preliminary experiments using Cohere's \textbf{Command R+} model,\footnote{\url{https://docs.cohere.com/docs/command-r-plus}} a 104B parameter instruction-tuned language model with 128K token context that is designed for multilingual tasks.

\subsection{Prompting}
\label{sec:prompts}
Though the exact prompt varies from experiment to experiment, all runs use the same base prompt. 


We use the following prompts for our preliminary experiments. The \textcolor{blue}{blue placeholders} are replaced with the appropriate values. The system prompt is as follows.
\begin{lstlisting}
You are an expert documentary linguist, specializing in (*@\textcolor{blue}{\$language}@*). You are working on a documentation project for (*@\textcolor{blue}{\$language}@*) text, where you are creating annotated text corpora using the interlinear glossed text (IGT) and following the Leipzig glossing conventions.

Specifically, you will be provided with a line of text in (*@\textcolor{blue}{\$language}@*) as well as a translation of the text into $metalang, in the following format.

Transcription: some text in (*@\textcolor{blue}{\$language}@*)
Translation: translation of the transcription line in (*@\textcolor{blue}{\$metalang}@*)

You are to output the gloss line of IGT. You should gloss stem/lexical morphemes with their translation in (*@\textcolor{blue}{\$metalang}@*), and gloss gram/functional morphemes with a label indicating their function. Please output the gloss line in the following format:

Glosses: the gloss line for the transcribed text

Glosses should use all caps lettering for functional morphemes and standard lettering for stem translations. Glosses for morphemes in a word should be separated by dashes, and words should be separated by spaces.
\end{lstlisting}

\newpage
The main prompt is as follows:

\begin{lstlisting}
Here are some complete glossed examples:
(*@\textcolor{blue}{\$fewshot\_examples}@*)

Please gloss the following example in (*@\textcolor{blue}{\$metalang}@*).

Transcription: (*@\textcolor{blue}{\$transcription}@*)
Translation: (*@\textcolor{blue}{\$translation}@*)
\end{lstlisting}

For zero-shot prompts, we remove the first sentence of the main prompt. Furthermore, from qualitative analysis, we observe that the LLM sometimes pulls words from the translation to use as glosses, resulting in incorrect examples. Thus, for the final test, we omit the translation lines from both prompts.

We run each experiment three times with temperature 0 and a different random seed, ensuring both the retrieval strategy and model API calls are reproducible. We report the average and standard deviation for performance.

\begin{figure*}[!bt]
  \centering
  
  \begin{minipage}[b]{0.24\linewidth}
    \centering
    \includegraphics[width=\linewidth]{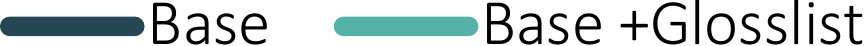} 
  \end{minipage}
  
  \vspace{1em} 

  \begin{minipage}[b]{0.24\linewidth}
    \includegraphics[width=\linewidth]{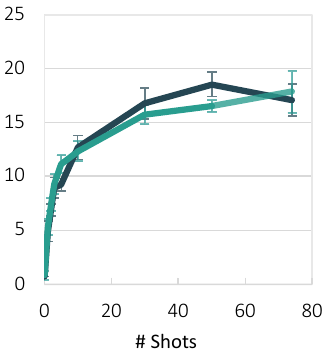}
    \subcaption{Gitksan}
    \label{fig:shots-git}
  \end{minipage}
  \hfill
  \begin{minipage}[b]{0.24\linewidth}
    \includegraphics[width=\linewidth]{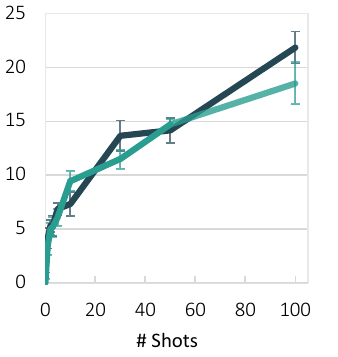}
    \subcaption{Lezgi}
    \label{fig:shots-lez}
  \end{minipage}
  \hfill
  \begin{minipage}[b]{0.24\linewidth}
    \includegraphics[width=\linewidth]{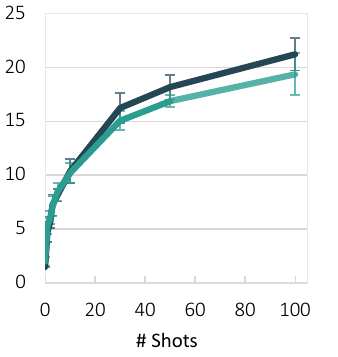}
    \subcaption{Natugu}
    \label{fig:shots-ntu}
  \end{minipage}
  \hfill
  \begin{minipage}[b]{0.24\linewidth}
    \includegraphics[width=\linewidth]{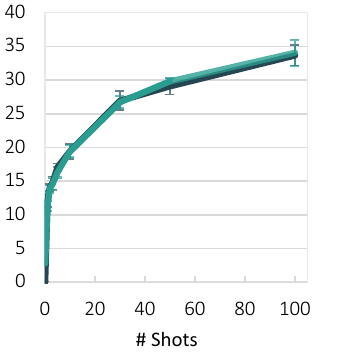}
    \subcaption{Uspanteko}
    \label{fig:shots-usp}
  \end{minipage}

\caption{Morpheme accuracy of LLM-based glossing on Gitksan, Lezgi, Natugu, Uspanteko, and [New Image Caption], varying the number of provided examples. Reported values are averages over three runs; error bars indicate standard deviation. In the \textsc{Base +Glosslist} setting, we provide a list of possible glosses in the prompt.}
\label{fig:shots}
\end{figure*}

\section{Many-Shot Prompting}
\label{sec:num-shots}

Few-shot prompting, where a model is provided with a small number of examples in the context, has proven very effective at a variety of tasks \citep{brown2020fewshot, winata-etal-2021-language, lin-etal-2022-shot, cahyawijaya2024llms}. Furthermore, as model context lengths have continued to increase, it has become possible to provide hundreds or even thousands of examples, and performance typically continues to improve \citep{bertsch2024incontext}. On the other hand, increasingly long prompts bear a high cost, and strategies to retrieve relevant examples can often achieve similar performance at a fraction of the cost (see \autoref{sec:retrieval}).

\subsection{Experimental Settings}
For all experiments, we run two settings, one with just the base task description, and one where we include a list of possible glosses for functional morphemes. We scrape this list of glosses from all of the seen glosses in the training set. We instruct the model to only use these glosses for functional morphemes (while stem morphemes should still be glossed with their translation). We refer to this setting as \textsc{[+ Glosslist]}, with an example gloss list in \autoref{sec:glosslist}.

For each language, we experiment with varying number of examples. For all languages except Gitksan, we run experiments providing no examples (zero-shot) and 1, 2, 3, 5, 10, 30, 50, and 100 examples. Gitksan has fewer than 100 training examples, so we use all 74 for the final setting.

For each example in our eval set, we randomly sample examples from the training set to be included in the prompt. In \autoref{sec:retrieval}, we compare this strategy to more intentional retrieval strategies that aim to select relevant examples.

\subsection{Results}

We report results for our languages in \autoref{fig:shots}, with a full table of results provided in \autoref{sec:full-results}. Generally, we see that the model has very weak performance in the zero-shot setting, indicating that the model has little knowledge of our chosen languages. In some cases, the zeroshot experiments produce results that are not even in the desired output format. 

Performance improves drastically for the first few shots added, showing smaller improvements as the number of shots increases. For Gitksan, performance levels up as the number of provided examples approaches the full training set. For the other languages with much larger training sets, performance shows continued improvement even around 100 shots, supporting the findings of \citet{bertsch2024incontext}. We suspect that this trend would continue to some extent, but the cost of providing hundreds of examples quickly becomes infeasible.

\begin{figure}[htb]
    \centering
    \includegraphics[width=\linewidth]{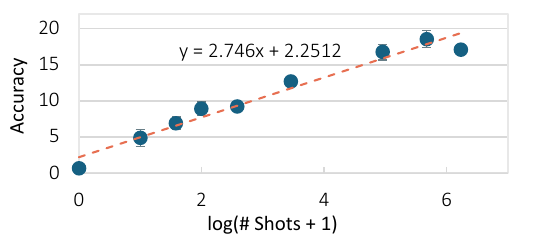}
    \caption{Morpheme accuracy for Gitksan, where the predictor variable is the logarithm of the number of provided examples (plus one).}
    \label{fig:logshots}
\end{figure}

\paragraph{Relationship between Shots and Accuracy}
What sort of shape is formed by the curve in \autoref{fig:gitx-shots} and \autoref{fig:shots}? The relationship appears to be roughly logarithmic, starting steep and leveling off. To quantify this relationship, we take the $\log(\# shots+1)$ for each setting.\footnote{Adding 1 so the zero-shot setting is defined.} \autoref{fig:logshots} shows the transformed curve for Gitksan, which now shows a strong linear relationship. 

We compute the $R^2$ value over all settings and report it in \autoref{tab:logcurvefits}.

\begin{table}[h]
    \centering
    \begin{tabular}{c|c c}
    \hline 
        Language & Base & + Glosslist  \\
        \hline
        Gitksan & 0.962 & 0.958\\
        Lezgi & 0.934 & 0.981 \\
        Natugu & 0.993  & 0.996 \\
        Uspanteko & 0.952 & 0.983 \\
         \hline 
    \end{tabular}
    \caption{Coefficient of determination ($R^2$) computed between morpheme accuracy and $\log(\# shots + 1)$}.
    \label{tab:logcurvefits}
\end{table}

We observe extremely strong correlation values across all settings. This indicates that the logarithmic model is a good fit for the data, and predicts that maintaining steady performance improvements requires exponentially more examples.

\begin{figure}[hbt]
    \centering
    \includegraphics[width=\linewidth]{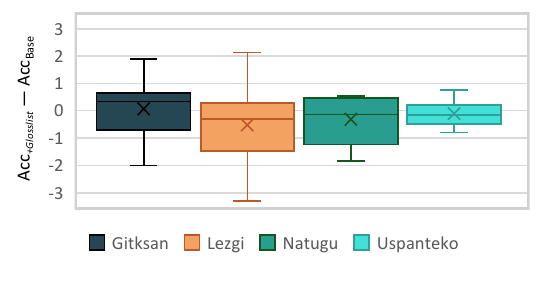}
    \caption{Difference in averaged accuracy between settings with and without a complete gloss list provided in the prompt. We observe minimal differences.}
    \label{fig:glosslist}
\end{figure}

\paragraph{Effect of Gloss List}
We initially hypothesized that providing a complete list of possible glosses in the prompt could help the model better adhere to the desired glossing conventions. We report a summary plot of the difference in accuracy between the two settings across languages in \autoref{fig:glosslist}.

The average difference is close to 0, well within a standard deviation in all cases, and thus there is little evidence to suggest that including the gloss list meaningfully affects performance. A possible explanation is that since the model has very limited prior knowledge of these languages, providing a simple list of glosses without any explanation or examples does not provide any useful information. 

\begin{figure}[htb]
    \centering
    \includegraphics[width=\linewidth]{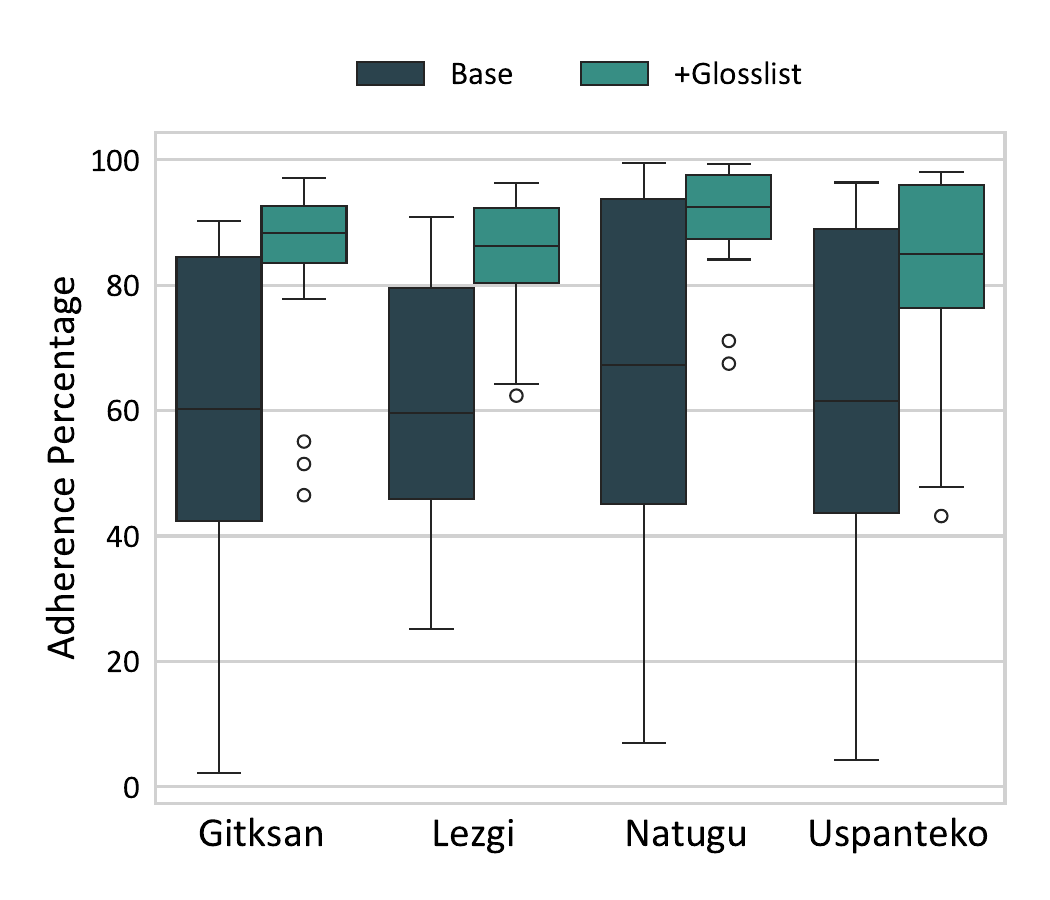}
    \caption{Distribution of adherence percentages, across languages, comparing with and without the glosslist.}
    \label{fig:adherence}
\end{figure}

To investigate whether including a gloss list changes the predictions at all, even if it doesn't improve glossing performance, we measure the \textit{adherence percentage}.
This metric is computed by dividing the number of predicted (functional) glosses that adhere to the gloss list by the total number of predicted glosses. We report the distribution over languages and settings in \autoref{fig:adherence}.

We observe that including the gloss list in the prompt is effective for increasing adherence compared to the base setting. While the experiments without the gloss list vary widely, the experiments with it nearly always use glosses from the list. On the other hand, we have observed no evidence that the gloss list improves performance, suggesting that the model may be predicting glosses from the list randomly. 


Furthermore, including a gloss list in the prompt carries a fixed cost of several hundred tokens for every prompt (e.g. for Uspanteko, the cost is 124 tokens). Since it provides negligible benefit, we opt to omit the glosslist for future experiments in order to reduce cost.

\section{Retrieval Strategies}
\label{sec:retrieval}
While including a large number of in-context examples can certainly improve performance, long prompts carry a high cost that may be infeasible for real-world documentation projects. For example, running prompts with a thousand examples in Uspanteko costs roughly 10 cents per inference call, which can quickly add up over thousands of examples. Many LLMs still have limited context length, particularly among open-source models, and including many examples may not even be possible. Finally, \citet{bertsch2024incontext} suggests that the effectiveness of many-shot prompting is mainly due to the model seeing relevant examples, and ignoring many irrelevant ones.

With this in mind, we consider a method inspired by \textbf{retrieval-augmented generation} (RAG, \citealt{lewis2020rag}). RAG was originally used for knowledge-intensive tasks, using document embeddings to search for relevant documents to a given query and include them in prompt context. We apply a similar strategy in order to search for relevant IGT examples from our training corpus to include in our prompt.

\subsection{Experimental Settings}
We consider several strategies for selecting examples that are relevant for the target sentence.

\paragraph{Random} As a baseline, we use the random strategy from the prior section, which simply samples $n$ examples randomly from the training corpus. 

\paragraph{Word Recall and Word Precision} We hypothesize that a straightforward way to improve performance is by providing examples which have the same morphemes as the target sentence. Since our data is not segmented into morphemes, we instead look for matching words (which will nearly always be composed of the same morphemes). We split each example into words using whitespace, and compute the \textit{word recall} for a target sentence $T$ and candidate training sentence $S$.
\begin{equation}
    \textsc{WordRecall} = \frac{|\textnormal{unique}(S) \cap \textnormal{unique}(T)|}{|\textnormal{unique}(T)|}
\end{equation}

This computes the fraction of unique words in the target sentence that appear in the candidate sentence. We can also compute the \textit{word precision} with a slightly modified formula:
\begin{equation}
    \textsc{WordPrecision} = \frac{|S \cap \textnormal{unique}(T)|}{|S|}
\end{equation}

This metric rewards examples where the majority of words in the candidate are in the target sentence. Notice that we do not use the unique words of $S$, instead weighting an example that uses the same word from $T$ several times more heavily. We select the examples with the highest word recall or precision, considering each example independently and breaking ties randomly.

\paragraph{Aggregate Word Recall}
One limitation of the prior approach is that by considering each candidate individually, we can potentially select several redundent examples in few-shot scenarios. Instead, we can compute the \textit{aggregate word recall} over a candidate sample of $n$ examples.
\begin{gather}
    S_{agg} = \bigcup_{i=1}^n \textnormal{unique}(S_i) \\
    \textsc{AggWordRec} = \frac{|S_{agg} \cap \textnormal{unique}(T)|}{|\textnormal{unique}(T)|}
\end{gather}

This metric rewards samples that jointly cover more of the words in the target. This is equivalent to the \textit{Maximum Coverage Problem}, and as such is NP-Hard \citep{nemhauser1978analysis}. We use the greedy algorithm, which runs in polynomial time \citep{hochbaum1996approximating}.

\paragraph{chrF}
A limitation of the previous strategies is that, by only considering atomic words, there is no way to select examples that may contain the same morphological units. One way we can attempt to capture morphological similarity is through using substring similarity metrics such as \texttt{chrF} \citep{popovic-2015-chrf} and \texttt{chrF++} \citep{popovic-2017-chrf}. These metrics compute the F-score of character n-gram matches (\texttt{chrF++} also incorporates word n-grams), and have been shown to correspond more closely to human judgements for machine translation. 

\paragraph{Morpheme Recall}
Although we do not have segmented data, much research has explored methods to induce morphological segmentations from data in an unsupervised manner. In particular, we use Morfessor \citep{Creutz2005}, a popular statistical method that seeks to find a segmentation that maximizes overall the probability of segmented words. 

We create silver segmentations using Morfessor and compute the recall metric as described earlier, but using morphemes rather than words. We train the segmentation model using the default parameters on the training data, and use Viterbi inference to segment test examples. We use the Morfessor 2.0 library \citep{virpioja2013morfessor}.

\subsection{Results}
\begin{figure*}[!bt]
  \centering
    \includegraphics[width=0.95\linewidth]{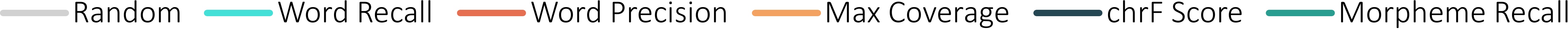}
  \\
  \vspace{10pt}
  \begin{minipage}[b]{0.5\linewidth}
    \includegraphics[width=\linewidth]{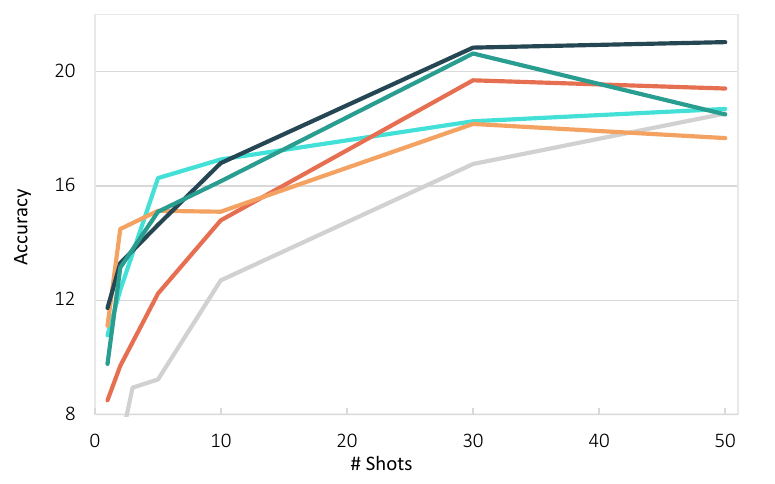}
    \subcaption{Gitksan}
    \label{fig:retrieval-git}
  \end{minipage}
  \hfill
  \begin{minipage}[b]{0.48\linewidth}
    \includegraphics[width=\linewidth]{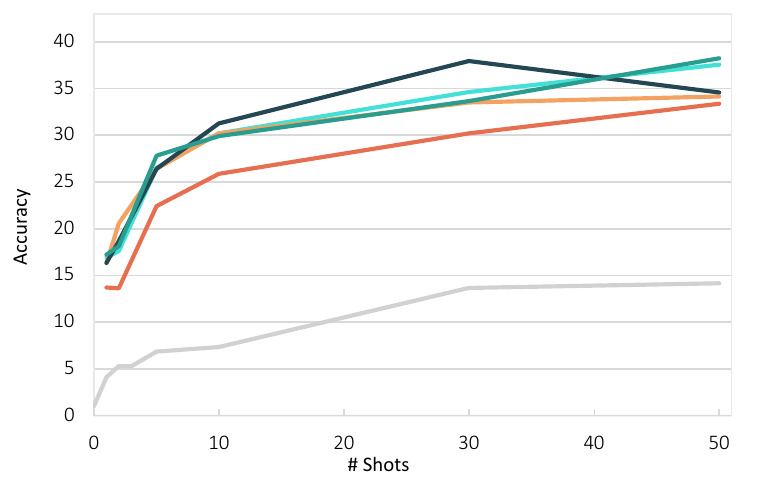}
    \subcaption{Lezgi}
    \label{fig:retrieval-lez}
  \end{minipage}
  \\
  \vspace{10pt}
  \begin{minipage}[b]{0.48\linewidth}
    \includegraphics[width=\linewidth]{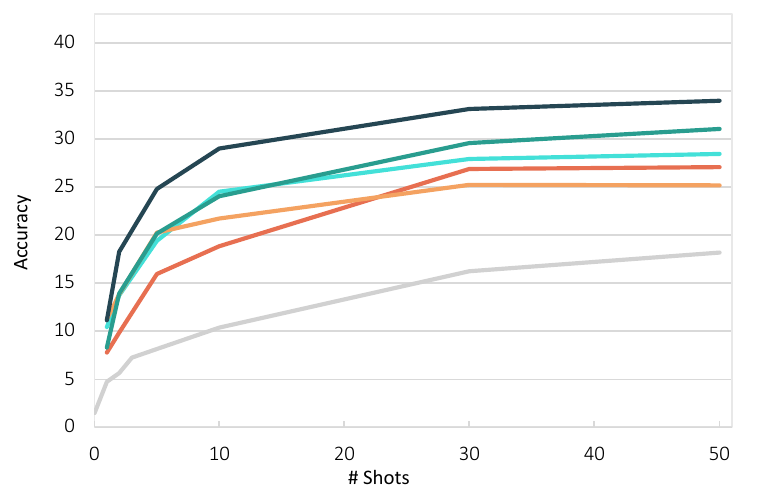}
    \subcaption{Natugu}
    \label{fig:retrieval-ntu}
  \end{minipage}
  \hfill
  \begin{minipage}[b]{0.48\linewidth}
    \includegraphics[width=\linewidth]{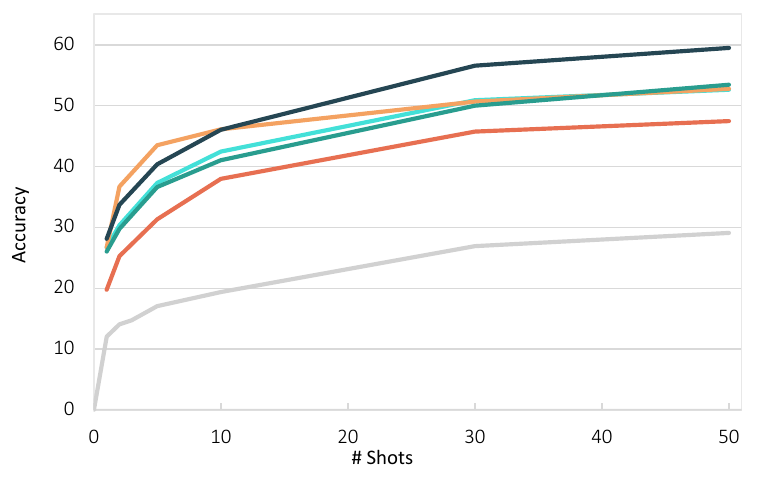}
    \subcaption{Uspanteko}
    \label{fig:retrieval-usp}
  \end{minipage}

\caption{Morpheme accuracy of LLM-based glossing on four languages, varying the number of provided examples and using different strategies to select relevant examples. Reported values are averages over three runs.}
    \label{fig:retrieval_results}
\end{figure*}

We report results across our four languages and six retrieval strategies in \autoref{fig:retrieval_results}. We run tests using 1, 2, 5, 10, 30, and 50 examples in each prompt.

\paragraph{Comparison with Random Retrieval} Across all languages, we observe clear and significant improvements over the random selection method described in the prior section (here indicated with a gray line). This is the case both with a small number of fewshot examples and as the number grows large. The only exception is the 50 example setting for Gitksan, at which point the provided examples make up a large fraction of the training corpus.

This is an intuitive result, as the IGT generation task requires, at minimum, knowledge about the words of a language and their potential glosses. Even a simple baseline that glosses tokens with their most common gloss from the training set is often fairly effective \citep{ginn-etal-2023-findings}. This is particularly important since the LLM used seems to have very limited prior knowledge of the language, as evidenced by the poor zero-shot performance.

\paragraph{Relationship between Shots and Accuracy} As before, we generally see consistently improving performance as additional examples are added. However, there are several cases where performance drops going from 30 to 50 shots, as in Gitksan (Word Precision, Max Coverage, and Morpheme Recall) and Lezgi (chrF Score). Both of these languages have fairly small corpora, and it is possible that after a point these strategies run out of beneficial examples, and any additional examples simply contribute noise to the prompt.

\paragraph{Effect of Different Granularities} Many of the strategies perform very similarly, but there are some observable trends across granularity levels (word, morpheme, and substring). We observe that the chrF strategy is nearly always the most effective, outperforming the word- and morpheme-based strategies in most cases. We hypothesize that this strategy strikes a balance by selecting examples with subword similarity, but not introducing error due to noisy morpheme segmentations. 

\paragraph{Word Recall vs Morpheme Recall} We observe mixed results across the Word Recall and Morpheme Recall strategies. We observe a few settings where there appears to be a significant gap between the two (Gitksan at 30 shots; Lezgi at 50 shots), but generally the strategies are close. It is possible that the words in our evaluation examples often either are monomorphemic, or contain a combination of morphemes already observed in the training data, and thus selecting relevant examples according to morphemes has little benefit.

\paragraph{Word Recall vs Word Precision}
While the Word Recall and Word Precision strategies both seek to quantify the word-level similarity between the target and candidate sentences, they are computed slightly differently and produce different results. The Word Recall strategy prioritizes candidate sentences that contain a large fraction of the word \textit{types} in the target sentence, ignoring repeated words. Meanwhile, the Word Precision strategy selects candidates based on the fraction of words within the candidate that are also in the target. 

The Word Recall strategy consistently outperforms Word Precision, except for the two largest settings in Gitksan. This indicates that it is more important to provide examples which cover the words in the target than it is to provide several examples for a single word. 

\paragraph{Word Recall vs Max Word Coverage} We experimented with the Max Word Coverage setting, where we consider the recall of the selected set of candidates as a whole, rather than individually. We observe minimal benefits, in fact underperforming the Word Recall setting in many cases.

\section{Comparison with SOTA}
Finally, we compare our best-performing strategies from the prior section with several previous baseline methods:
\begin{itemize}
    \setlength\itemsep{0em}
    \item The \textbf{token classification} transformer model of \citet{ginn-etal-2023-findings}, which uses an encoder model to predict glosses word-by-word
    \item \textbf{T\"{u}-CL} from \citet{girrbach-2023-tu-cl}, which uses hard attention to induce latent segmentations and predict glosses on segmented words
\end{itemize}
\noindent
For the LLM-based method, we select the chrF strategy and test with 30 examples for Gitksan and 100 examples for the other languages. We make some small prompt optimizations described in \autoref{sec:prompts}, and raise the temperature to 0.2. We use the following language models:
\begin{itemize}
    \setlength\itemsep{0em}
    \item Cohere's \textbf{Command R+}, which was used for preliminary experiments.
    \item OpenAI's \textbf{GPT-4o}, specifically the \texttt{gpt- 4o-2024-05-13} checkpoint \citep{openai2024gpt4}
    \item Meta's \textbf{Llama 3.1} 8b parameter model \citep{dubey2024llama3herdmodels}, using the 8-bit quantization and the MLX \citep{mlx2023} checkpoint.
    \item Google's \textbf{Gemini 1.5 Pro} \citep{geminiteam2024gemini}
\end{itemize}

\begin{figure}
    \centering
    \includegraphics[width=\linewidth]{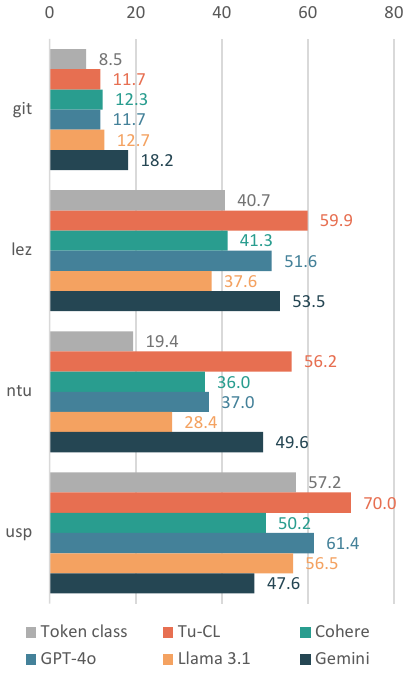}
    \caption{Morpheme accuracy results on test splits, comparing several LLMs and baseline systems.}
    \label{fig:test-results}
\end{figure}

\noindent
We run evaluation on the held out test set and report results in \autoref{fig:test-results}.

\subsection{Discussion}
We observe that the LLM based glossing strategies outperform a simple transformer in nearly every setting, despite using no training whatsoever and using a small fraction of the training set as examples. Even the Llama 8b parameter model, an open-source model that can be run on a laptop, is competitive.

Of the LLM models, Gemini performs best on three languages. However, we note that Gemini refuses to produce answers for many examples, which we count as completely wrong. If we omit such examples, Gemini's performance is even higher, achieving 55.9\%, 50.8\%, and 63.9\% accuracy on Lezgi, Natugu, and Uspanteko respectively.

On the other hand, the LLM methods typically underperform the SOTA method of \citet{girrbach-2023-tu-cl}, except for Gitksan, where the best LLM (Gemini) outperforms by 6.5 points. The \citet{girrbach-2023-tu-cl} approach explicitly models segmentation through a learned latent representation, which our strategy does not utilize. Future work with LLM-based methods could explore an analogous process, explicitly prompting the LLM to generate segmentations before producing final glosses.

Furthermore, these methods will likely continue to improve as LLMs become more capable for rare (or even completely unseen) languages, as measured by benchmarks such as \citet{tanzer2024benchmark}. Most trivially, as LLMs with increasingly long contexts are developed, we can provide more examples in-context, which our results indicate will continue to provide benefits.

\section{Conclusion}
We find that SOTA large language models struggle to produce interlinear glosses for the endangered languages used in our research. However, by selecting relevant examples from a training corpus and providing them as part of the context for each example to be glossed, we can significantly improve performance. We find that the relationship between performance and the number of few-shot examples is roughly logarithmic. Performance improves 
by a wide margin when we select examples with a high chrF++ score relative to the target sentence.

Our best systems outperform a standard transformer model, despite involving no explicit training and using a fraction of the training data. However, they still underperform the SOTA system for the glossing task on three out of four languages. Thus, for documentary linguists hoping to use automated glossing solutions, the use of LLMs may not achieve ideal accuracy.
At the same time, LLMs may still be a preferrable choice for languages with very limited data comparable to Gitksan, and the use of an API is often far more accessible than training and hosting a neural model. Our results encourage further exploration of this approach.

\section*{Limitations}
While we have selected a small set of languages that we believe give insight into the performance of automated glossing systems, they are certainly not representative of all the world's languages. In particular, LLMs may struggle more with languages that use non-Latin writing scripts \citep{NEURIPS2023_117c5c86}.

We use a single prompt template for the majority of experiments and do not conduct extensive prompt engineering. Frameworks such as DSPy \citep{khattab2024dspy} have shown that prompt optimization can often greatly improve performance, so it is entirely possible that we could achieve better performance on this problem with the same models and strategies.

We evaluate three popular closed-source LLMs, and one smaller open-source LLM, but results may vary across other models. 

\section*{Ethics Statement}
As our work involves documentation data produced through the combined efforts of documentary linguists and speakers of endangered languages, we strive to respect their desires and avoid treating data as merely a resource to train models with \citep{schwartz-2022-primum}.

We do not intend for automated glossing systems to replace human annotators, which would drastically impact the quality, novelty, and utility of annotated corpora, but rather to serve as a tool available to support documenters. 

Finally, we acknowledge that the use of large language models carries a high environmental cost, and make efforts to minimize unnecessary API calls and to track our usage.

\section*{Acknowledgements}
We thank the anonymous reviewers for their helpful comments. We also thank the Input Experience team at Apple for their feedback on a preliminary version of this work.

This work was supported by compute credits from a Cohere For AI Research Grant, these grants are designed to support academic partners conducting research with the goal of releasing scientific artifacts and data for good projects. 

Parts of this work were supported by the National Science
Foundation under Grant No. 2149404, “CAREER: From One Language to Another.” Any opinions, findings, and conclusions or recommendations expressed in this material are those of the authors and do not necessarily reflect the views of the National
Science Foundation.

\bibliography{anthology, custom}

\appendix
\newpage
\section{Example Gloss List}
\label{sec:glosslist}
We provide an example list of glosses for Gitksan. There are some formatting artificats, due to the automatic extraction of glosses.
\begin{lstlisting}
#(PROSP), (#COMP), (#PROSP), 1.I, 1.SG.=, 1PL.II, 1SG, 1SG.II, 2SG, 3.I, 3.II, 3.III, 3PL, 3PL.II, 3PL.INDP, 3SG.II, ANTIP, AX, CAUS1, CAUS2, CCNJ, CN, CNTR, COMP, CONNN, DEM.PROX, DES, DISTR, DM, DWID, EPIS, FOC, FUT, FUT=3, IBM, INCEP, INS, IPFV, IPFV=EPIS=CN, IRR, IRR=3, LOC, LOC=CN, LVB, MANR, NEG, NEG=FOC, NEG=FOC=3, NMLZ, OBL, PART, PASS, PCNJ, PN, PR.EVID, PREP, PREP=CN, PROG=CN, PROG[=CN], PROSP, PROSP=3, PROSP=3.I, REAS, SELF, SG, SPT, SX, T, T=PN, TR, TR=CN, TR=PN, VAL, VER, VERUM, [#(PROSP), [(#COMP), [(PROSP), [PROG=CN, [PROSP
\end{lstlisting}

We chose to provide just the list of glosses, without any additional information, to replicate the scenario where there are no additional resources other than glossed examples. Of course, if we had access to a dictionary or grammar reference, providing this information could be beneficial.

\section{Full Results}
\label{sec:full-results}
We present full results across all of our experimental settings in \autoref{tab:results}.

\begin{table*}[p]
\small
    \centering
    \begin{tabular}{l|c c c c c c c c c}
    \toprule
         & \multicolumn{9}{c}{\textbf{\# In-Context Examples}}\\ \\
         
       \textbf{Strategy} & 0 & 1 & 2 & 3 & 5 & 10 & 30 & 50 & 100 \\
        \midrule
        \multicolumn{10}{l}{\textbf{\textsc{Gitksan}}} \\
        \midrule
        Random & 0.7±0.0 & 4.9±1.2 & 6.9±0.9 & 8.9±0.9 & 9.2±0.4 & 12.7±0.7 & 16.8±1.1 & 18.5±1.2 & 17.1±0.3 \\
        Rand \textsc{+Gloss} & 0.8±0.2 & 5.3±0.9 & 7.4±1.0 & 9.3±1.1 & 11.1±1.8 & 12.3±0.8 & 15.7±1.3 & 16.5±1.7 & 17.9±0.7 \\
        Word Rec. & & 10.8±0.3 & 12.4±0.9 & & 16.3±4.6 & 16.9±1.1 & 18.3±1.4 & 18.7±1.3 &\\
        Word Prec. & & 8.5±0.3 & 9.7±0.5 & & 12.2±0.6 & 14.8±1.1 & 19.7±0.3 & 19.4±0.5 & \\
        MaxWordCov. & & 11.1±1.9 & 14.5±1.7 & & 15.1±0.8 & 15.1±0.5 & 18.2±1.5 & 17.7±0.3 &  \\
        chrF & & 11.7±0.4 & 13.3±0.3 & & 14.6±1.1 & 16.8±0.8 & 20.8±0.4 & 21.0±0.6 &\\
        Morph. Rec. & & 9.8±0.2 & 13.1±0.5 & & 15.1±0.7 & 16.2±1.2 & 20.6±2.2 & 18.5±0.7 &\\
        
        \midrule
        \multicolumn{10}{l}{\textbf{\textsc{Lezgi}}} \\
        \midrule
        Random & 1.0±0.2 & 4.1±0.6 & 5.3±0.6 & 5.3±0.8 & 6.9±1.6 & 7.3±0.6 & 13.7±1.2 & 14.2±1.4 & 21.8±6.0 \\
        Rand \textsc{+Gloss} & 1.0±0.1 & 3.4±0.1 & 5.0±0.7 & 5.2±1.0 & 6.1±0.7 & 9.5±0.7 & 11.5±1.6 & 14.7±3.8 & 18.5±0.1 \\
        Word Rec. & & 17.0±0.7 & 17.6±2.8 & & 26.5±1.5 & 30.2±2.1 & 34.6±1.6 & 37.6±1.5 &\\
        Word Prec. & &  13.7±1.3 & 13.6±0.8 && 22.4±1.6 & 25.9±1.4 & 30.2±1.7 & 33.4±1.9 & \\
        MaxWordCov. & & 16.3±0.4 & 20.6±2.6 & & 26.4±0.9 & 30.2±1.3 & 33.5±1.2 & 34.1±1.4 & \\
        chrF & & 16.4±1.6 & 18.7±0.5 & & 26.4±0.8 & 31.3±0.7 & 37.9±0.4 & 34.6±1.1 &\\
        Morph. Rec. & & 17.2±0.9 & 18.1±0.5 & & 27.8±0.1 & 29.9±3.4 & 33.6±1.3 & 38.2±1.9 &\\
        \midrule
        \multicolumn{10}{l}{\textbf{\textsc{Natugu}}} \\
        \midrule
        Random & 1.5±0.3 & 4.7±0.4 & 5.6±0.3 & 7.2±0.7 & 8.1±0.7 & 10.4±0.3 & 16.2±1.3 & 18.2±1.4 & 21.2±0.3 \\
        Rand \textsc{+Gloss} & 2.0±0.2 & 5.3±0.4 & 6.1±0.4 & 7.1±1.0 & 8.4±0.3 & 10.2±0.7 & 15.1±1.4 & 16.9±1.0 & 19.4±0.6 \\
        Word Rec. & & 10.4±0.4 & 13.7±0.6 & & 19.4±1.0 & 24.5±1.8 & 27.9±1.6 & 28.4±2.1 & \\
        Word Prec. & & 7.8±0.2 & 9.9±0.5 & & 16.0±0.2 & 18.8±1.5 & 26.9±0.8 & 27.0±1.0 & \\
        MaxWordCov. & &  11.2±0.3 & 13.8±0.3 & & 20.2±0.3 & 21.7±1.0 & 25.2±2.2 & 25.2±1.0 & \\
        chrF & & 11.1±0.4 & 18.2±0.7 & & 24.8±0.5 & 29.0±1.4 & 33.1±0.9 & 34.0±0.5 & \\
        Morph. Rec. & & 8.3±0.5 & 13.9±0.3 & & 20.2±2.0 & 24.0±1.9 & 29.6±1.9 & 31.0±1.4 & \\
        \midrule
        \multicolumn{10}{l}{\textbf{\textsc{Uspanteko}}} \\
        \midrule
        Random & 2.7±0.3 & 12.1±0.9 & 14.1±0.6 & 14.7±1.0 & 17.1±0.6 & 19.4±1.1 & 26.9±1.4 & 29.1±1.2 & 33.7±1.5 \\
        Rand \textsc{+Gloss} & 2.8±0.4 & 11.3±0.8 & 13.9±0.6 & 14.6±0.9 & 16.3±0.8 & 19.4±0.9 & 26.7±0.9 & 29.8±0.5 & 34.1±1.9 \\
        Word Rec. & & 26.7±1.4 & 30.4±1.6 & & 37.3±1.3 & 42.4±0.8 & 50.9±0.2 & 52.6±0.7 & \\
        Word Prec. & & 19.7±0.2 & 25.3±0.4 & & 31.3±1.0 & 37.9±0.6 & 45.7±0.4 & 47.5±0.8 &  \\
        MaxWordCov. & & 26.7±1.2 & 36.7±1.0 & & 43.5±1.7 & 46.1±1.0 & 50.7±2.2 & 52.8±2.0 & \\
        chrF & & 28.1±0.7 & 33.7±0.7 & & 40.4±0.1 & 46.0±0.2 & 56.5±0.7 & 59.5±0.7 & \\
        Morph. Rec. & & 26.1±0.7 & 29.8±0.8 &  &36.6±0.1 & 41.0±1.3 & 50.0±0.4 & 53.4±0.3 & \\
        \bottomrule
    \end{tabular}
    \caption{Full morpheme accuracy results across languages, selection strategies, and number of examples. \textsc{+Gloss} indicates the gloss list was included in the prompt.}
    \label{tab:results}
\end{table*}

\end{document}